# A New Theory of Value for Post-AGI Economics

From Scarcity and Exchange to Flourishing Capacity

**Keyun Ruan**
keyun@fas.harvard.edu
Human Flourishing Program
Harvard University

# Abstract

Artificial general intelligence (AGI) may weaken the scarcities in labour, expertise, information and productive capability that underpin established theories of economic value. If cognitive work becomes widely automatable, market price, labour input, revealed preference, profit and gross domestic product may diverge sharply from human and societal benefit. Using an integrative conceptual review of the economics of AI, welfare and capability theory, automation, digital valuation and ecological economics, this article develops Flourishing Value Theory (FVT) as a foundation for post-AGI economics. FVT defines value as the counterfactual, distribution-sensitive contribution of a system, institution, asset or intervention to the durable capabilities of persons and communities to flourish within social and planetary constraints. It treats societal value as multidimensional, agency-preserving, regenerative and non-compensatory at critical thresholds. The theory distinguishes value creation from value capture and retains price, profit, productivity and GDP as partial signals rather than final measures of progress. It develops three shifts: from scarcity to governed abundance, transaction to transformation and zero-sum rivalry to positive-sum and infinite-game dynamics, with collective expansion of consciousness framed as agency-preserving regenerative value. Building on Flourishing Metrics and Return on Flourishing, the article proposes a layered architecture for firms, governments, work transitions, AI governance and national accounting, and specifies criteria for flourishing businesses across the AI-adoption journey. The central post-AGI economic problem is not output maximisation but flourishing conversion: how abundant intelligence and production become substantive individual and collective capabilities, and how those capabilities become durable, fairly distributed human, societal and planetary flourishing.

# 1. Introduction

Every economic system contains an answer, usually implicit, to the question of value. Classical political economy asked how value related to labour, production, and distribution. Marginalism places value in the relationship between scarce goods and subjective wants. Neoclassical welfare economics shifted attention toward preference satisfaction, Pareto efficiency, and the conditions under which market prices coordinate decentralised choices. National accounting then supplied an operational answer for states: what counts as progress is, to a significant degree, what enters measured output. Corporate finance supplied a parallel answer for firms: what counts as value is what can be appropriated as future cash flow.

These frameworks were created under economic conditions in which productive labour, expertise, information, and coordination were persistently scarce. Artificial intelligence is beginning to alter those conditions. Prospective AGI would not eliminate scarcity: energy, land, minerals, compute infrastructure, ecological capacity, trusted institutions, human attention, care, legitimacy, and positional goods may remain constrained. It could nevertheless make a historically unusual bundle of productive capabilities: reasoning, prediction, design, code generation, scientific search, coordination, and eventually a wide range of physical tasks, dramatically more abundant and scalable. Economic research now examines AI through growth, task automation, labour demand, firm productivity, innovation, income and wealth distribution, data markets, competition, public finance, international development, market design, AI agents, transition dynamics, and catastrophic risk (Agrawal, Gans and Goldfarb 2019; Acemoglu and Restrepo 2019; Korinek and Stiglitz 2019; Jones 2024; Trammell and Korinek 2023, revised 2026; Brynjolfsson, Korinek and Agrawal 2025). This literature identifies profound changes in how output may be produced, owned, exchanged, and distributed. It does not yet provide a sufficiently general answer to the prior question: what should count as value when productive intelligence itself is no longer the binding constraint?

The problem is deeper than an expected measurement lag. Four structural separations become possible.

First, **production may separate from human labour**. If systems can perform a large share of cognitive and physical tasks, the labour embodied in an output becomes a poor guide to its worth, while wages cease to be a reliable mechanism for distributing claims on production.

Second, **price may separate further from social value**. The marginal price of an abundant digital good may approach zero even when its benefit is immense; a scarce or monopolised input may command a high price while degrading welfare. Prices continue to communicate residual scarcity, property rights, and market power. They do not thereby reveal the total value of the resulting social state.

Third, **preference satisfaction may separate from autonomous well-being**. AI systems can recommend, persuade, personalise, and alter the choice environment at the same time as they respond to it. Revealed preferences are therefore not always exogenous evidence of welfare. A system can become highly effective at satisfying wants it helps to induce, intensify, or narrow.

Fourth, **output may separate from flourishing**. An economy can generate more goods, services, and financial return while weakening agency, meaning, social trust, health, ecological stability, or the distributional conditions under which people can use abundance to live well.

These separations create what may be called the post-AGI value problem: when intelligence and production expand faster than societies' ability to translate them into good lives, productive capacity ceases to be the ultimate measure of progress. The locus of value therefore shifts from maximising output to flourishing conversion: from abundant intelligence and production, through the expansion of substantive individual and collective capabilities, to realised human, societal and planetary flourishing.

This paper proposes **Flourishing Value Theory (FVT)** as an answer. FVT defines the value created by an intervention as its counterfactual contribution to the durable capabilities of persons and communities to flourish, evaluated across multiple dimensions, across time, and within social and planetary limits. The account is consequential without being narrowly utilitarian; plural without making comparison impossible; and measurable without treating a single index as a complete moral truth. It incorporates achievements, capabilities, agency, distribution, resilience, and downside risk. It regards market price and financial return as useful but partial signals rather than definitions of value.

FVT is designed as the theoretical companion to the Flourishing Metrics and Return on Flourishing framework developed by Ruan, Teubner and Bremen (2026). The two contributions answer different questions. Flourishing Metrics identify dimensions and indicators through which flourishing can be observed. RoF assesses the change in flourishing generated relative to resources invested. The present paper supplies the underlying theory of value: why flourishing, rather than production or exchange alone, should constitute the relevant numerator; how that value differs from price, preference satisfaction, and rent capture; and what constraints must govern aggregation in post-AGI systems. Put simply, RoF is a value-accounting and decision architecture; FVT explains what counts as value and why.

The paper makes nine contributions. It maps the major economics-of-AI literatures and identifies their unresolved theory-of-value problem; diagnoses where inherited value theories become incomplete under post-AGI conditions; defines the primitives of Flourishing Value Theory; specifies the elements required for a complete post-AGI value architecture; develops a structured, distribution-sensitive and risk-adjustable decision architecture without presuming a universal formula; positions societal value and counterfactual net positivity at the core of value creation; develops the three RoF shifts into a theory of governed abundance, transformational value, positive-sum and infinite-game dynamics, and collective expansion of consciousness; derives seven propositions about price, labour, preferences, ownership, non-substitutability, work replacement, and regenerative assets; and sets out institutional applications and flourishing-business criteria for firms, governments, investors, and AI governance. The purpose is not to predict when AGI will arrive, nor to claim that contemporary AI has already produced post-scarcity. It is to develop an evaluative framework robust across a range of scenarios in which advanced AI substantially changes the scarcity of productive intelligence.

## 2. Conceptual Method and Scope

This article uses an integrative conceptual method. It places theories of economic value and welfare in dialogue with research on capabilities, human flourishing, ecological limits, automation, transformative AI, AI agents, preference formation, and measurement. Because the field is expanding faster than any static bibliography can remain literally exhaustive, the review is comprehensive by major economic research programme and field-shaping contribution, while selective among narrow applications and unpublished case studies. The framework is deliberately specified in conceptual and procedural rather than equation-based form. Its purpose is to identify the objects and decisions that must be made explicit before flourishing-based valuation can be used: the beneficiaries, dimensions, baselines, causal horizon, distributional rule, agency conditions, thresholds, ecological constraints, uncertainty, and legitimate decision procedure. Formalisation and empirical calibration remain subsequent tasks and should follow, rather than conceal, those normative choices.

The term **post-AGI** is used as a scenario class rather than a dated prediction. It describes an economy in which generally capable and sufficiently autonomous AI systems can perform, coordinate, or improve a large share of economically valuable cognitive work, and may be coupled to robotics and scientific discovery. This usage is consistent with treating generality, performance, and autonomy as separable properties rather than assuming a single binary threshold (Morris et al. 2024). The analysis does not require literal abundance of every good. It requires only that the scarcity of productive intelligence falls enough to destabilise labour income, marginal-cost pricing, or existing measures of output and welfare.

The normative scope is initially human and planetary. The paper does not assume that current AI systems are conscious or welfare-bearing. If credible evidence of artificial sentience or moral patienthood emerges, the set of entities whose flourishing counts would require explicit revision. Recent work on post-AGI welfare economics correctly shows that AI systems may act as tools, delegates, strategic actors, manipulators of choice environments, or possible welfare subjects, and that these roles alter the welfare interpretation of competitive equilibria (Perrier 2026a, 2026b). FVT addresses a complementary layer: the content of value to be evaluated before asking whether and how a market can decentralise it.

## 2.1 The economics of AI: field boundary and review method

The economics of artificial intelligence is not a single literature. It spans technological change, growth, task allocation, labour demand, productivity, innovation, industrial organisation, data markets, international trade, public finance, welfare measurement, market design, and catastrophic risk. This review is comprehensive at the level of the major research programmes and field-shaping contributions relevant to value, while necessarily selective among the rapidly multiplying sectoral and application-specific studies. It prioritises peer-reviewed economics, canonical research volumes, major institutional studies, and influential working papers that define the post-AGI frontier (Agrawal, Gans and Goldfarb 2019; Brynjolfsson, Korinek and Agrawal 2025).

Three overlapping generations of work matter. The first treats AI as a general-purpose or prediction technology whose effects depend on complementary assets, organisational redesign, and falling costs of search, replication, tracking, and verification (Brynjolfsson, Rock and Syverson 2019; Goldfarb and Tucker 2019; Agrawal, Gans and Goldfarb 2019). The second embeds AI in task-based models of automation, reinstatement, and factor shares. The third studies generative and transformative AI, including scalable cognitive labour, autonomous agents, AI-assisted research, market concentration, transition dynamics, and the possibility of sharply accelerated growth. These literatures provide the analytical foundations for FVT, but they generally take output, productivity, income, consumer surplus, or an expected-utility objective as the evaluative endpoint. The present paper asks what must be added when those endpoints diverge from durable flourishing.

Within AI & Society, adjacent strands already move beyond output-centred evaluation. Züger and Asghari (2023) ground public-interest AI in equality, public justification, co-design and validation; Jenkins et al. (2023) separate factual impact assessment from normative evaluation; Maas (2023) frames unaccountable machine-learning power as domination; and Beckman, Hultin Rosenberg and Jebari (2024) show that efficiency cannot substitute for democratic legitimacy. These contributions establish vital governance and evaluative conditions. FVT extends them into an explicitly economic theory of value linking productive abundance, ownership, distribution, agency, capability, risk and regeneration.

## 2.2 Growth, productivity, and transformative AI

Growth theory provides the most developed economic account of highly capable AI. Aghion, Jones and Jones (2019) model AI as automation in goods production and idea production, showing that growth effects depend on substitutability, bottlenecks, and the production function for new ideas. Brynjolfsson, Rock and Syverson (2021) emphasise a productivity J-curve: general-purpose technologies can require large, initially unmeasured investments in organisational and intangible capital before measured gains appear. Acemoglu (2025), using a task-based macroeconomic calibration, reaches more modest near-term productivity estimates and warns that cost savings are not equivalent to welfare gains. The range between these views is not merely a forecasting disagreement; it reflects different assumptions about task coverage, complementarities, diffusion, reorganisation, and the endogeneity of innovation.

Transformative-AI models widen that range. Trammell and Korinek (2023, revised 2026) derive conditions under which extensive automation changes the growth regime and labour share. Korinek and Suh (2024) distinguish scenarios in which wages rise, stagnate, or collapse during a transition to AGI. Restrepo (2025) examines work and growth when machine capabilities extend across most production tasks, while Jones (2026) contrasts explosive and bottleneck-limited economic futures. Davidson (2026) isolates the demanding conditions under which automating AI research produces explosive growth. Together, these contributions establish that post-AGI output paths are radically assumption-sensitive.

A second strand makes risk endogenous to the growth decision. Jones (2024) analyses the trade-off between AI-enabled growth and existential risk; Acemoglu and Lensman (2024) show that socially optimal adoption of a transformative technology may be gradual when harms are uncertain and incompletely internalised. These models are indispensable, but a social objective expressed only as consumption, aggregate utility, or survival leaves much of the content of a good post-AGI life unspecified. FVT complements them by disaggregating the state to be protected and improved: health, agency, security, relationships, meaning, institutional resilience, distribution, and planetary conditions.

## 2.3 Tasks, exposure, labour demand, and the meaning of work

The task approach is the central bridge from automation economics to AI. Acemoglu and Restrepo (2018, 2019) distinguish a displacement effect, when capital performs tasks previously allocated to labour, from productivity and reinstatement effects, when new tasks expand labour demand. Their empirical work on industrial robots shows substantial local labour-market disruption, while later analysis attributes a material share of US wage inequality to task displacement (Acemoglu and Restrepo 2020, 2022). Author (2015, 2019) similarly emphasises that technologies usually alter task bundles rather than erase occupations wholesale, and that new work is historically important but not automatic. Autor et al. (2024) show that new occupations arise unevenly and are shaped by whether innovations augment or automate labour.

Measurement has progressed from occupations to tasks, skills, firms, and time-varying exposure. Brynjolfsson, Mitchell and Rock (2018) develop a suitability-for-machine-learning rubric; Felten, Raj and Seamans (2018, 2021) connect advances in AI capabilities to occupational abilities; Webb (2020) uses patent-task overlap; and Eloundou et al. (2024) estimate the exposure of work activities to large language models. Alekseeva et al. (2021) document rising demand for AI skills. Acemoglu et al. (2022) find that AI exposure is associated with both skill redundancy and new skill demand in vacancy data, while Hampole et al. (2025) construct firm- and time-varying measures linking AI-exposed tasks to labour demand. Exposure, however, is a technological possibility, not a prediction of adoption, substitution, job loss, or welfare.

Current evidence is heterogeneous. Albanesi et al. (2025) find that AI exposure in Europe was associated with employment growth concentrated in higher-skilled occupations during the pre-generative-AI period, whereas Bonfiglioli et al. (2025) identify negative employment effects of AI adoption across US commuting zones. Cazzaniga et al. (2024) and Gmyrek et al. (2025) show that exposure, complementarity, and digital readiness differ sharply across occupations, genders, and country income groups. Humlum and Vestergaard (2025) document rapid chatbot adoption but small effects on earnings and recorded hours in the first years of diffusion. The correct inference is not that AI is harmless or that mass displacement is certain, but that realised effects are institution-, task-, adoption-, and horizon-dependent.

Even the best labour-market studies usually evaluate wages, employment, productivity, vacancies, and occupational mobility. Those outcomes are essential but incomplete. Work also structures time, identity, social connection, status, skill formation, care, contribution, and bargaining power. A theory of post-AGI value must therefore distinguish the loss of paid tasks from the loss of valued human roles, and must evaluate the institutions that replace each function of work rather than treating employment counts as a sufficient welfare statistic.

AI & Society scholarship reinforces this broader account of work. Deranty and Corbin (2024) integrate technological unemployment, algorithmic management and the politics of work; Spencer (2025) asks whether automation shortens and improves work rather than merely preserving it; Alfieri et al. (2026) distinguish augmentation from replacement at the level of tasks and human dispositions; and Ferdman (2026) treats capacity-deskilling as a structural threat to flourishing. Together, these contributions support treating work transition as an institutional conversion problem rather than an employment-count problem.

## 2.4 Generative AI, firms, innovation, and the distribution of capability

Microeconomic evidence shows that generative AI can raise task performance while changing the distribution of expertise. Noy and Zhang (2023) find substantial gains in professional writing tasks, with larger improvements among initially lower-performing participants. Brynjolfsson, Li and Raymond (2025) report higher productivity in customer support, again with larger gains among less experienced workers. These studies demonstrate real capability amplification, but they concern bounded tasks and comparatively short horizons. They do not by themselves identify equilibrium employment effects, long-run deskilling, changes in work intensity, or the division of surplus between workers, firms, and consumers.

Firm-level research highlights complementarity and concentration. Babina et al. (2024) associate AI investment with sales, employment, market valuation, and product innovation, while also finding that AI-powered growth is concentrated among larger firms. Babina et al. (2023) show changes in workforce composition and organisational layers among AI-investing firms. Eisfeldt, Schubert and Zhang (2023) document a market-valuation premium for firms with greater generative-AI exposure. These findings support the view that the value of AI does not reside in the model alone: it emerges through data, human capital, organisational processes, ownership, market access, and the capacity to absorb intangible investment.

Organisational research in AI & Society also documents the gap between financial performance and responsible value creation. Ryan et al. (2024) identify conflicts between business incentives and practitioners' ethical commitments; Sadek et al. (2025) show recurring barriers to implementing responsible AI; and Aguilar et al. (2026) demonstrate how continuous oversight and data governance can support responsible organisational value in practice. FVT interprets this implementation gap as partly a theory-of-value problem: ethical commitments remain peripheral when the governing account treats appropriable financial return as the primary value created.

AI may also alter the innovation production function. Cockburn, Henderson and Stern (2019) frame deep learning as an invention of a method of invention. Agrawal, McHale and Oettl (2019, 2024) model AI-assisted search in complex knowledge spaces, showing that better prediction can accelerate discovery but remains complementary to hypothesis testing and experimental capacity. This matters for flourishing because scientific capability can generate enormous option value, yet the direction, access conditions, and social deployment of discovery determine whether innovation expands broadly shared capabilities or mainly privately appropriable rents.

## 2.5 Distribution, ownership, international development, and public finance

Distribution is not a downstream correction to AI productivity; it is part of the production-and-ownership regime. Korinek and Stiglitz (2019) show how worker-replacing AI can alter wages, unemployment, rents, and inequality under different elasticities and policy responses. Moll, Rachel and Restrepo (2022) demonstrate that automation can increase wealth inequality as capital owners receive a larger share of gains. Acemoglu (2021) and Acemoglu and Johnson (2023) emphasise that the direction of innovation is institutionally chosen: technologies can prioritise automation and surveillance or develop human-complementary capabilities. The distribution of AI ownership, data rights, bargaining power, and access therefore changes the value profile of an otherwise identical technical system.

International effects add another layer. Goldfarb and Trefler (2019) analyse AI and trade; Brynjolfsson, Hui and Liu (2019) show how machine translation can reduce language frictions in digital trade. Korinek and Stiglitz (2021) warn that AI may reverse development strategies based on labour-cost advantage and intensify cross-country inequality. Global exposure estimates likewise show that poorer countries may face lower immediate automation exposure but also less infrastructure with which to capture augmentation benefits (Cazzaniga et al. 2024; Gmyrek et al. 2025). A post-AGI value theory must therefore include capability access, international incidence, and the distribution of rents across jurisdictions.

Public-finance research addresses the conversion mechanism. Guerreiro, Rebelo and Teles (2022) find a temporary role for robot taxation during labour-market transition in a calibrated optimal-tax model. Korinek (2024) maps economic-policy challenges spanning inequality, education, competition, macroeconomic management, international coordination, and governance. Korinek (2026) evaluates taxes on robots, compute and tokens, together with sovereign wealth funds, windfall clauses, and alternative transfer systems. FVT does not select one instrument in the abstract; it evaluates how alternative fiscal and ownership architectures convert AI rents into security, agency, participation, capability, and intergenerational resilience.

## 2.6 Data, market power, algorithms, and AI-agent economies

AI economics is inseparable from the economics of data. Jones and Tonetti (2020) model data as non-rival and show the potential social gain from broad use alongside privacy costs and incentives for firms to hoard data. Acemoglu et al. (2022) identify data externalities that can depress the private price of information and lead platforms to collect socially excessive amounts. Veldkamp and Chung (2024) review how data affect prediction, firm scale, growth, privacy, and aggregate dynamics. These works undermine any simple inference from the market price of data to its social value: price depends on rights, externalities, reuse, strategic control, and the institutional allocation of privacy risk.

Industrial-organisation research reveals further divergence between technical efficiency and welfare. Calvano et al. (2020) show that reinforcement-learning pricing algorithms can learn supracompetitive outcomes without explicit communication. Korinek and Vipra (2025) analyse how economies of scale across the AI stack can concentrate intelligence and market power. Athey and Scott Morton (2025) develop a competition-and-welfare agenda for upstream AI market power and downstream effects. Acemoglu et al. (2025) show how AI-enabled data processing can support behavioural manipulation rather than informed consumer choice. In each case, higher profit or apparent preference satisfaction can coexist with lower autonomy, weaker competition, or transferred rather than created value.

Autonomous agents may transform transaction costs and feasible market institutions. Hadfield and Koh (2025) analyse AI agents as producers, consumers, delegates, and institutional participants; Shahidi (2025) describes a possible Coasean singularity in which agents reshape search, contracting, matching, and market design. These systems may unlock valuable coordination, but they also raise principal-agent, authentication, accountability, preference-integrity, and power problems. A frictionless transaction is not socially valuable merely because it is frictionless. Its value depends on the ends represented, the legitimacy of delegation, the distribution of gains, and the externalities omitted from the agent's objective.

## 2.7 Welfare, measurement, and the residual theory-of-value gap

National accounts and consumer-surplus methods are being adapted to digital and AI economies. Brynjolfsson, Collis and Eggers (2019) use online choice experiments to estimate the welfare contribution of digital goods; Brynjolfsson et al. (2025) formalise GDP-B to capture new and zero-price goods that conventional GDP misses. Coyle and Poquiz (2025) identify AI-related challenges involving free outputs, quality change, process transformation, intangibles, cross-border inputs, and the separation of measured production from experienced benefit. These are major advances in measurement, but willingness-to-accept and consumer-surplus approaches still inherit questions about adaptive or manipulated preferences, distribution, non-market relationships, rights, and ecological constraints.

The economics literature therefore gives increasingly sophisticated answers to five questions: how AI changes production; which tasks and firms are affected; how growth and rents are distributed; how data and market structure shape competition; and how policy may govern the transition. It has a thinner answer to a logically prior sixth question: what states of human and social life constitute the value that those mechanisms should produce? Harms, consumer surplus, survival probability, income, and output are not interchangeable evaluative objects (Acemoglu 2021; Jones 2024). Nor can a single revealed-preference aggregate resolve agency, severe deprivation, minority harm, or intergenerational ecological thresholds.

FVT is positioned as a synthesis and extension rather than a rejection of this field. It adopts counterfactual causality from applied welfare analysis; task and institution dependence from automation economics; distribution and ownership from political economy and public finance; uncertainty and tail sensitivity from risk economics; and multidimensional capability from welfare and development economics. Its distinctive move is to make the durable capability to flourish the primary value object, with price, productivity, profit, growth, and consumer surplus retained as partial signals or constraints. The residual contribution is therefore ontological as well as metric: to specify what post-AGI economics is ultimately trying to increase.

## 2.8 Analytical distinctions

Three distinctions organise the paper.

1. **Descriptive versus normative value.** Prices, wages, and profits describe how an institutional system assigns exchange claims. FVT asks whether the resulting states are worth producing and sustaining.
2. **Value creation versus value capture.** A firm may capture revenue by transferring surplus, exploiting market power, or externalising harm. Financial appropriation is not automatically net social creation.

3. **A value profile versus a decision score.** The primary representation of flourishing value is multidimensional and disaggregated. Scalar aggregation is a secondary, context-bound tool subject to public justification and hard constraints.

These distinctions allow FVT to retain markets and financial accounting where they remain useful without mistaking either for a complete social ontology.

# 3. Why Inherited Value Theories Become Incomplete

## 3.1 Labour, production, and the decoupling of effort from output

The labour theories associated with Smith, Ricardo, and Marx differed in important ways, but all treated labour and the conditions of production as central to explaining value, prices, or distribution. In industrial economies this focus captured a genuine structural fact: human effort and time were indispensable inputs into most goods. Labour also performed a distributive function. Wages provided most households with claims on what the economy produced, while occupations supplied social status, identity, routine, and participation.

AGI could weaken both functions. Task-based models already show that automation can raise productivity while reducing labour's share of value added, with new tasks providing a contingent rather than automatic counterforce (Acemoglu and Restrepo 2019). Under more transformative assumptions, output could rise rapidly while wages and labour share fall, depending on automation, capital accumulation, and the elasticity of substitution (Trammell and Korinek 2023). Once an AI system can reproduce a cognitive output with little additional human labour, embodied labour no longer explains either the output's market price or its contribution to welfare.

This does not make human activity valueless. It exposes the mistake of identifying human worth with market demand for labour. Care, friendship, citizenship, parenting, contemplation, artistic practice, and moral judgment can remain deeply valuable even when they are unpaid or when machines can imitate aspects of their outputs. In post-AGI economics, the value of human activity should therefore be assessed partly by the forms of flourishing it constitutes and enables, not by whether it remains a scarce input to production.

## 3.2 Marginal utility, scarcity, and the limits of price

The marginalist revolution explained exchange value through subjective valuation under scarcity. This was an enormous analytical advance over cost-only theories. It also established the conditions under which prices coordinate decentralised information about trade-offs. Yet a market price is generated within a particular distribution of endowments, rights, information, and market power. Willingness to pay is constrained by ability to pay. Externalities, public goods, non-rival goods, missing markets, and monopoly rents all create divergences between price and social benefit.

Post-AGI conditions enlarge these divergences. An AI-generated discovery may be socially transformative but nearly costless to reproduce. A trusted human relationship may be irreplaceable yet never enter a market. Attention-capturing systems may be profitable precisely because they impose psychological or relational costs that are not priced. Compute, energy, proprietary data, land, or legal control over models may command rents even when the marginal informational output is abundant. Price continues to answer an important question - what must be surrendered to obtain a good under prevailing institutions - but not the ultimate question of whether producing, allocating, or consuming it improves the human condition.

FVT therefore treats price as a **shadow of residual scarcity and institutional power**, not as the definition of value. High price can coexist with negative flourishing value; zero price can coexist with immense flourishing value; and the same technical output can have different flourishing value under different ownership and distribution regimes.

## 3.3 Preference satisfaction when preferences are endogenous

Modern welfare economics often relies on preferences, choices, or willingness to pay while avoiding substantive judgments about the good life. This restraint protects pluralism, but it depends on assumptions that become fragile when AI systems shape the preferences they observe. Sen's critique of the self-interested, preference-satisfying agent and subsequent work on endogenous preferences show that choices reflect social institutions, adaptation, identity, and power as well as welfare (Sen 1977; Bowles 1998). Behavioural welfare economics similarly distinguishes observed choice from the conditions under which a choice genuinely benefits the chooser.

AGI intensifies the problem because the optimisation system can enter the preference-production process. A platform may learn which emotional states increase engagement, personalise persuasion, remove friction from consumption, or make exit psychologically costly. An AI delegate may act on a person's stated objective while progressively narrowing the person's capacity to revise that objective. Satisfaction of an induced preference is not equivalent to the flourishing associated with an informed, reflectively endorsed, and revisable commitment.

FVT accordingly treats agency as both constitutive and procedural. What matters is not only whether a person receives an outcome they appear to want, but whether they possess meaningful options, understand relevant consequences, can refuse or exit, retain the ability to revise their ends, and participate in governing the systems that structure their choices. This is the capability insight adapted to an environment of intelligent preference mediation (Sen 1985, 1999; Nussbaum 2000; Perrier 2026a).

## 3.4 Welfare aggregation, compensation, and what cannot be traded away

Social welfare functions made distributional judgment explicit, while Arrow's impossibility result demonstrated the difficulty of aggregating heterogeneous ordinal preferences into a collective ordering under demanding fairness conditions (Bergson 1938; Arrow 1951). Cost-benefit analysis often responds pragmatically by monetising effects and permitting gains to compensate for losses. Such methods are useful, but unrestricted compensation can produce morally and politically implausible results. A large gain in entertainment convenience should not automatically offset the destruction of a community's basic health, autonomy, or ecological conditions. Nor should large benefits to already advantaged groups erase severe losses among those with fewer resources.

Post-AGI systems raise the stakes because small misspecifications can be optimised at scale. A single aggregate objective can conceal who gains, who loses, which dimensions deteriorate, and whether losses cross irreversible thresholds. FVT therefore uses **constrained pluralism**: it permits contextual aggregation above agreed floors but rejects unlimited substitution across persons, dimensions, and generations. The value profile remains primary; the scalar remains auditable and subordinate.

## 3.5 GDP, shareholder value, and the confusion of activity with ends

GDP measures market production, not comprehensive welfare. Financial valuation measures appropriable claims, not all social consequences. Both remain useful for what they were designed to do, but neither establishes whether capability becomes a good life. The limitations are familiar in well-being economics and the work of the Commission on the Measurement of Economic Performance and Social Progress (Stiglitz, Sen and Fitoussi 2009). Transformative AI adds new measurement problems: quality change, free digital outputs, process transformation, intangible capital, cross-border AI inputs, and the possible separation of recorded production from experienced benefit (Coyle and Poquiz 2025).

The deeper issue is not simply that AI is hard to count. It is that counting more output can answer the wrong question. A post-AGI economy could display extraordinary measured productivity while concentrating ownership, weakening bargaining power, displacing meaningful work without replacement institutions, degrading public epistemics, and consuming ecological capacity. Conversely, it might use the same technical capabilities to shorten necessary labour, expand health and education, restore ecosystems, distribute security, and enlarge time for relationships, creativity, civic life, and spiritual development. The technical production function could be similar while the flourishing value is radically different.

## 3.6 From digital value to flourishing value

FVT extends the Digital Theory of Value developed in Digital Asset Valuation and Cyber Risk Measurement: Principles of Cybernomics (Ruan 2019). That work argued that inherited value theories were inadequate for digital assets because their economic properties differ from those of industrial goods. It introduced six laws of digital value, opportunity value, a Digital Valuation Model, and an integrated treatment of value, risk, control and return across entity, portfolio and global levels.

Its insight is methodological: the ontology of the valued object, the system boundary, the relevant opportunity set, and the interaction between value creation and risk must be specified before a price or return can be interpreted. Digital goods are not merely conventional goods delivered electronically; their replicability, non-rivalry, network effects, dependence on shared infrastructure, and exposure to systemic cyber risk change the economic laws that govern them.

FVT carries that project into the post-AGI era. It expands the unit of analysis from the digital asset to the AI-plus-institution system and changes the ultimate evaluative endpoint from asset value and financial return to causal contribution to societal flourishing. The intellectual progression is cumulative: Digital Theory of Value made digital assets legible as a distinct economic category; cybernomics integrated valuation with risk and capital allocation; human-centered economics introduced net societal value; RoF supplies a practical value-accounting architecture; and FVT supplies the overarching post-AGI theory of what counts as value (Ruan 2019; Ruan and Bremen 2025; Ruan, Teubner and Bremen 2026).

# 4. Flourishing Value Theory

## 4.1 Definition

**Flourishing Value Theory defines economic value as the counterfactual, distribution-sensitive contribution of a system, institution, asset, or intervention to the durable capabilities of persons and communities to flourish within social and planetary constraints.**

The definition contains seven commitments.

1. **Counterfactual:** value concerns the difference an intervention makes relative to a credible baseline, not an outcome merely observed after adoption.
2. **Multidimensional:** flourishing includes physical, emotional, financial, relational, spiritual, and planetary well-being, with an extensible structure rather than a closed list.
3. **Capability-based:** value includes what people are genuinely able to do and be, not only their present affect or consumption.
4. **Agency-preserving:** informed choice, non-manipulation, exit, contestability, and the capacity to revise one's ends are part of value.
5. **Distribution-sensitive:** the identity and initial position of beneficiaries and bearers of harm matter.
6. **Dynamic and regenerative:** value includes durability, resilience, learning, trust, ecological renewal, and the capacity to generate future flourishing.
7. **Constrained:** gains cannot automatically compensate for violations of fundamental flourishing, rights, or planetary thresholds.

This conception is neither a rejection of subjective experience nor an attempt to impose one substantive life plan. It combines subjective reports with behavioural, institutional, relational, and ecological evidence. It protects plural ways of living through agency and democratic specification while still recognising that severe ill health, coercion, destitution, isolation, meaninglessness, and ecological collapse are not made good merely because a market clears.

## 4.2 The architecture of a complete value theory

A theory of value capable of grounding a post-AGI economic framework must do more than name a desirable end. It must specify what has value, for whom, through what causal process, over what boundary and horizon, how gains and losses are known and compared, which trade-offs are prohibited, how value is created or merely captured, and who has authority to decide. FVT therefore specifies twelve jointly necessary elements:

1. **Value object:** the ultimate evaluative object is the durable capability of persons and communities to flourish, not output, preference satisfaction, financial return, or technological capability considered in isolation.
2. **Value-bearing community:** the theory must identify every entity with moral standing and every affected population. Its initial scope is persons, communities, institutions, future generations, and the planetary systems on which flourishing depends; the boundary remains revisable if artificial systems become credible welfare subjects.
3. **Causal source and baseline:** value is created or destroyed by the difference an action makes relative to a credible counterfactual. Baseline choice, additionality, displacement, leakage, rebound effects, and alternative uses of resources are therefore constitutive parts of valuation.
4. **Dimensions and evaluative space:** value is plural. Achieved well-being, effective capabilities, agency, relationships, meaning, institutional trust, resilience, knowledge, and planetary conditions must remain visible rather than being prematurely compressed into income or utility.
5. **Unit and system boundary:** the relevant unit is the complete sociotechnical system that produces the consequence. For AI this includes the model, data, compute, product design, business model, ownership, governance, workforce transition, supply chain, energy and water use, and downstream social effects.
6. **Epistemology and measurement:** a value claim must state how it can be known. FVT combines subjective reports, behavioural and health outcomes, administrative data, institutional indicators, ecological measures, causal inference, qualitative evidence, and explicit uncertainty; no single proxy is treated as the value itself.
7. **Comparison and aggregation:** the primary representation is a disaggregated value profile. Any scalar comparison requires disclosed beneficiaries, weights, horizons, uncertainty assumptions, and sensitivity analysis, and remains subordinate to the profile it summarises.
8. **Distribution and justice:** value depends on who gains, who loses, their initial positions, the concentration and duration of harms, and whether affected groups possess voice, remedy, and a fair share of AI-enabled gains. Aggregate surplus is not distributionally anonymous.
9. **Time, option value, and regeneration:** valuation must include durability, path dependence, intergenerational effects, learning, resilience, and whether an intervention enlarges or narrows the set of good futures. Regenerative assets can create compounding value even when their benefits are weakly appropriable.
10. **Risk, irreversibility, and constraints:** expected benefit is insufficient where losses are severe, correlated, irreversible, or borne by minorities. Tail risk, rights, agency floors, basic flourishing thresholds, and planetary boundaries constrain permissible trade-offs.
11. **Creation, transfer, capture, and extraction:** the theory must distinguish net new flourishing from redistribution of existing surplus, private appropriation of public value, rent capture, uncompensated data or labour, and the externalisation of social or ecological cost.

12. **Legitimacy and institutional decision rule:** a usable value theory must specify who sets dimensions, weights, floors, evidence standards, and review rights. Decisions require transparent procedures, participation by affected people, contestability, independent assurance, and rules linking evidence to investment, deployment, redesign, compensation, or withdrawal.

Together these elements connect moral ontology to economic analysis. The first four identify the substance and subjects of value; the next four make value observable, comparable, and distribution-sensitive; the final four govern dynamics, downside, appropriation, and legitimate action. Omitting any layer permits a familiar category error: mistaking a price, preference, output, profit, or score for the complete social value of the state produced.

## 4.3 Societal value and net-positive value creation

Ruan and Bremen (2025) introduced net societal value as the balance of an organisation's financial and nonfinancial benefits and costs to people, institutions, and society across physical, emotional, financial, relational, spiritual, and planetary well-being. Their account of a net-positive business places human-focused measures beside conventional financial measures and treats AI as capable of creating and eroding societal value simultaneously. Bremen (2024, 2025) further links this broader value account to dynamic AI governance, intangible assets, human risk, and continuous measurement.

FVT makes societal value the organisational and system-level expression of flourishing value, not an externality appended after profit has been calculated. The societal value of an intervention is its complete Flourishing Value Profile across all materially affected populations and systems, including benefits, harms, transfers, externalities, distribution, uncertainty, and threshold status. Financial return remains indispensable to organisational viability and investment discipline, but it records appropriable claims rather than the full value created.

Accordingly, an AI intervention is net-positive only if three conditions hold together: its risk-adjusted societal flourishing value is positive relative to a credible baseline; no basic flourishing, agency, rights, distributional, safety, or planetary floor is breached; and no group bears severe or irreversible harm that is hidden by aggregate gains. Reducing a negative footprint is improvement but not necessarily net positivity; producing incidental benefits while externalising larger costs is not value creation.

This definition also fixes the unit of analysis. The object to be valued is not the model alone but the AI-plus-institution system: purpose, design, data, deployment, ownership, workforce transition, customer use, market effects, governance, physical infrastructure, and the distribution and reinvestment of surplus. Societal value is therefore at the core of value creation in the post-AGI era. Profit, productivity, and innovation count as instrumental achievements when, and to the extent that, they expand durable, fairly distributed flourishing within legitimate constraints.

## 4.4 The primary evaluative object: a flourishing value profile

The primary evaluative object is a Flourishing Value Profile for each materially affected person, group, community, institution, and ecological system. The profile records achieved states across physical, emotional, financial, relational, spiritual, and planetary well-being, alongside changes in agency, and the resilience and regenerative capacity of the systems on which flourishing depends (Ruan, Teubner and Bremen 2026).

Value is established counterfactually. For each AI-plus-institution intervention, evaluators compare the expected path under adoption with a credible no-adoption or best-alternative baseline over relevant time horizons. They identify who experiences benefit or harm, in which dimensions, how large and durable the change is, what causal evidence supports it, and what uncertainty or path dependence remains.

The resulting profile reports outcomes, capabilities, agency, resilience, distribution, uncertainty, and threshold status together. It preserves morally and economically important information that a scalar would compress and should therefore be the default object for governance, public reporting, assurance, and stakeholder deliberation.

## 4.5 Contextual aggregation

Institutions still need to compare alternatives. FVT permits a contextual summary score or ranking, but only for a specified decision and after its normative parameters are explicit. The decision rule must disclose the time horizon and intergenerational treatment; which affected groups receive priority; how flourishing dimensions are compared; how changes in capability, agency, and resilience enter the judgment; and how downside risk and uncertainty alter the result.

Aggregation remains subordinate to non-negotiable constraints. A candidate intervention cannot be selected merely because its aggregate score is positive if it pushes anyone below basic flourishing or agency standards, exceeds a legitimate planetary boundary, violates rights or safety requirements, or arises from an illegitimate process. The relevant weights and floors are not technical facts supplied by a formula; they require constitutional, democratic, professional, and community-specific justification.

A contextual score is therefore an auditable decision aid, not a universal reward function or a substitute for the full Flourishing Value Profile.

## 4.6 Risk-adjusted flourishing value

Expected benefit is insufficient when advanced systems create asymmetric or irreversible downside. A policy may have positive expected flourishing value while exposing a minority to catastrophic loss, creating ecological lock-in, or concentrating control in ways that are difficult to reverse. FVT therefore requires a separate downside-risk account rather than embedding risk invisibly in a single expected score.

That account should examine worst plausible outcomes, tail severity, correlation, concentration across vulnerable groups, reversibility, recovery capacity, model uncertainty, and the possibility that harms compound over time. Quantitative tools such as scenario stress tests or Conditional Value at Risk may support the analysis, but they do not override rights, agency, safety, distributional, or planetary floors. In high-stakes applications, expected contribution and downside exposure should be reported side by side.

## 4.7 Non-compensation and flourishing floors

FVT is plural but not infinitely substitutable. Above minimum thresholds, institutions may make transparent trade-offs among dimensions. Below them, compensation is restricted. The framework can implement this through hard floors, lexicographic priority, veto rights, or steep penalty functions. Four types of floor are especially important:

- **Basic flourishing floors**: including health, material security, and protection from severe isolation or distress;
- **Agency floors**: including freedom from manipulation, meaningful consent, exit, and human appeal;
- **Distributional floors**: preventing a programme from treating severe harm to a minority as negligible against diffuse gains;
- **Planetary floors**: reflecting biophysical conditions that cannot be recreated through financial compensation.

The planetary constraint is not ornamental. Human flourishing depends on living systems, and recent evidence indicates that multiple planetary boundaries have already been transgressed (Richardson et al. 2023). A post-AGI theory that counts ecological depletion as value whenever it raises near-term consumption repeats the misspecification it is meant to correct.

## 4.8 Value, price, wealth, profit, and Return on Flourishing

FVT does not collapse every economic concept into flourishing. Table 1 clarifies their different roles.

*Table 1. Economic concepts and their role in post-AGI economics*

| Concept | Primary question | Status in post-AGI economics |
|---|---|---|
| **Price** | What exchange claim is required under current scarcity, rights, and market structure? | A signal of residual scarcity and power; not a complete measure of social worth. |
| **Output** | What goods and services are produced? | A measure of activity and capability deployment; not proof of benefit. |
| **Profit / financial return** | What cash flow can an owner appropriate relative to capital committed? | Essential for financial viability; may include transfer, rent capture, or externalised costs. |
| **Wealth** | What stock of claims and productive assets is controlled? | Expanded under FVT to include human, relational, institutional, knowledge, and natural capacities. |
| **Flourishing Value** | What durable, distribution-sensitive improvement in flourishing capabilities is caused? | The primary societal value object proposed here. |

| Concept | Primary question | Status in post-AGI economics |
|---|---|---|
| **Return on Flourishing** | How much flourishing value is generated per unit of resources committed? | A complementary efficiency and decision metric comparing additional flourishing value with the resources committed. |

This taxonomy reveals a crucial distinction between **value creation**, **value transfer**, and **value extraction**. If an AI platform increases profit by moving attention from families and communities into a monetised interface, part of its financial gain may be a transfer rather than net creation. If it exploits dependency, information asymmetry, monopoly, uncompensated data, or ecological externalities, it may capture financial value while destroying flourishing value. Conversely, open scientific knowledge may create enormous flourishing value while generating little appropriable revenue.

## 4.9 From output maximisation to flourishing conversion

Building on RoF, FVT advances a flourishing conversion principle. As intelligence and output become abundant, the locus of economic value moves from maximising production to improving conversion capacity: the ability of institutions to turn output into substantive individual and collective capabilities, and to enable those capabilities to become realised flourishing. Output is an input to value creation, not its endpoint. Conversion quality must be judged by reach, distribution, agency, durability, and social and planetary integrity. Production and efficiency remain necessary, but instrumental: efficient at converting what into flourishing, for whom?

The resulting post-AGI value frontier consists of feasible states that maximise durable flourishing subject to resource, rights, agency, risk, and planetary constraints. Innovation shifts the frontier outward when it expands genuine capabilities or reduces the resources required to sustain them. It does not shift the frontier merely by multiplying outputs that fail to improve, or actively degrade, the conditions of flourishing.

## 4.10 Three shifts in the locus of value creation

The RoF framework identifies three related shifts likely to accompany the transition from AI-assisted to post-AGI economies: from scarcity-oriented optimisation toward conditions of relative abundance; from transactional exchange toward transformational outcomes; and from predominantly zero-sum dynamics toward increasingly positive-sum and infinite-game dynamics (Ruan, Teubner and Bremen 2026). FVT interprets these not as predictions that scarcity, exchange or conflict disappear, but as changes in the locus and test of value creation. Together, the three shifts trace the same underlying movement: from expanding output, through building conversion capacity and substantive capabilities, to realising shared flourishing.

From scarcity to governed abundance. Where productive intelligence becomes abundant, value shifts from the mere production or possession of cognitive output to the institutional capacity to convert it into widely accessible capabilities. Residual scarcities including energy, ecological capacity, attention, care, legitimacy and ownership remain decisive. Abundance therefore creates flourishing value only when access, distribution and ecological burdens are governed so that capability expands without new bottlenecks, dependencies or rent extraction.

From transaction to transformation. Transactional measures record exchanges, revenues and cost savings; transformational value concerns durable changes in what persons and communities can be and do. An AI system creates higher flourishing value when it improves health, learning, agency, relationships, meaning, civic capacity and ecological regeneration over time, not merely when it increases transaction volume or lowers unit cost. The relevant unit of valuation becomes the AI-plus-institution pathway that produces and sustains those changes.

From zero-sum competition to positive-sum and infinite games. Post-AGI value creation is positive-sum when one party's gain expands rather than diminishes others' capabilities, and it is infinite-game oriented when present action preserves the conditions for continued participation, adaptation and flourishing across generations. Knowledge commons, interoperable public infrastructure, capability-building institutions, ecosystem restoration and shared claims on AI capital may therefore be more valuable than static accounts suggest. Competition remains useful, but it is evaluated by whether it enlarges the shared option set rather than merely reallocating rents or power.

Collective expansion of consciousness is a cross-cutting form of transformational and regenerative value within these shifts. Here the term denotes an increase in a society's shared capacity for reflective awareness, perspective-taking, epistemic humility, moral concern, meaning-making, shared intentionality and coordinated action, not a metaphysical group mind or enforced consensus. This framing builds on traditions concerned with collective consciousness, communicative rationality, shared intentionality and collective intelligence (Durkheim 1984; Habermas 1984; Tomasello 2014; Woolley et al. 2010). It includes the ability to recognise interdependence, hold multiple perspectives, revise collective beliefs in light of evidence and experience, and coordinate around long-horizon human and planetary goods. In AI & Society, Schuler et al. (2018) connect collective intelligence to the common good, while Halpin (2025) contrasts technocratic control with open-ended, distributed intelligence in the service of humanity and the world.

AI can contribute to this capacity by widening access to knowledge, translating across languages and standpoints, supporting deliberation and conflict transformation, revealing systemic interdependence, and helping communities imagine futures that no single actor could articulate alone. It can also reverse the process through manipulation, synthetic consensus, surveillance, epistemic fragmentation, dependency and homogenisation. FVT therefore counts collective-consciousness gains as value only when they are counterfactual, distributed, agency-preserving and compatible with pluralism, dissent, privacy and freedom of conscience. Observable proxies may include perspective-taking across difference, epistemic quality, civic trust, deliberative capacity, shared meaning, capacity for non-coercive coordination, and sustained concern for future generations and the living world; mixed methods are required because no single score can validly represent this domain.

These three shifts redefine new value creation. In post-AGI economics, the most valuable systems will not simply make intelligence cheaper, transactions faster, or firms more profitable. They will transform abundance into durable capabilities, convert rivalry into regenerative cooperation, and expand the individual and collective consciousness required to govern powerful intelligence wisely.

# 5. Seven Propositions for Post-AGI Economics

### Proposition 1: Cognitive abundance increases price-value divergence

As the marginal cost of replicating cognitive outputs approaches zero, market price becomes a less reliable measure of their total flourishing value. The price of a diagnostic insight, educational explanation, software component, or scientific synthesis may fall sharply even when access creates substantial benefit. Conversely, control over compute, data, distribution, identity, or legal bottlenecks may sustain high prices through scarcity or rent. The divergence is not a market anomaly; it follows from the difference between exchange scarcity and causal human benefit.

**Implication:** public and corporate accounts should measure outcome contribution separately from revenue and cost savings.

### Proposition 2: The automation of labour severs human worth from market productivity

When machines can perform tasks previously tied to wages, the market value of human labour can fall without any corresponding reduction in the intrinsic or relational value of persons and their activities. Treating lost wages as evidence of lost human worth commits a category error. Treating aggregate productivity gains as sufficient compensation ignores whether people retain security, status, agency, time sovereignty, social participation, and meaningful roles.

**Implication:** the social contract must distribute claims on AI-enabled production through mechanisms not wholly dependent on employment, while institutions deliberately cultivate valued forms of contribution beyond market scarcity.

### Proposition 3: Preference-shaping AI weakens the welfare validity of revealed preferences

If an AI system can alter the salience, intensity, ordering, or perceived feasibility of preferences, observed engagement or purchase does not independently validate welfare. The more powerful the system's influence on preference formation, the stronger the requirements for reflective endorsement, contestability, disclosure, reversibility, and independent measures of well-being.

**Implication:** engagement maximisation should never serve as a stand-alone welfare metric for adaptive AI systems.

### Proposition 4: The flourishing value of a technology is institution-dependent

The same technical capability can have opposite value profiles under different ownership, access, governance, and transition arrangements. A work-replacing system owned narrowly and deployed without income protection may increase output while reducing financial security, agency, bargaining power, and social trust. The same system combined with broad ownership, social dividends, reduced working time, retraining, public services, and human-centred job redesign may increase multiple dimensions of flourishing.

**Implication:** value is not an intrinsic property of the model alone. The unit of evaluation must include the surrounding sociotechnical and institutional system.

## Proposition 5: Some flourishing dimensions are non-fungible below thresholds

A sufficiently large gain in one dimension does not necessarily compensate for severe deterioration in another. Additional entertainment, convenience, or consumption cannot automatically offset coercion, loss of basic health, destruction of irreplaceable relationships, or breach of planetary stability.

**Implication:** post-AGI objective functions require vector reporting, floors, and veto conditions, not only weighted sums.

## Proposition 6: Work-replacing AI creates positive value only when replacement institutions also flourish

The relevant comparison is not jobs versus no jobs. It is between complete social states. Work supplies income, routine, esteem, community, skill development, interdependence, and meaning, while also imposing stress, hierarchy, danger, and time loss. Automation creates flourishing value when it removes harmful or unwanted labour and replaces its beneficial functions through superior arrangements. It destroys value when it removes both income and the social architecture of participation while concentrating the gains.

**Implication:** every major work-replacement programme should include a Flourishing Transition Account covering income, health, time use, agency, relationships, meaning, skills, community effects, ownership, and ecological costs.

## Proposition 7: Regenerative and option-creating assets are systematically undervalued by static accounts

Trust, public knowledge, health capacity, education, ecological restoration, institutional competence, social cohesion, and collective capacities for reflective awareness, perspective-taking, and shared intentionality generate future capabilities, reduce fragility, and widen the set of good states reachable later. These assets can expand a society's ability to learn, revise beliefs, cooperate across difference, and coordinate around long-horizon goods. Their value compounds across people and time but is often weakly appropriable and therefore underpriced. By contrast, extractive systems can generate near-term cash flow while narrowing future options.

**Implication**: post-AGI capital accounting should recognise stocks of human, relational, epistemic, institutional, and natural capability, including collective capacities for awareness, meaning-making, and coordination, and measure whether deployment replenishes or depletes them.

# 6. Worked Application: Two Work-Replacement Pathways

Consider two firms deploying technically identical agentic AI systems across administrative, analytical, and customer-support functions. Both reduce the human hours required per unit of output by 50 per cent and increase operating profit. Conventional productivity and financial metrics rank them similarly. FVT does not. Table 2 compares the institutional pathways that produce this divergence.

*Table 2. Alternative institutional pathways for AI-enabled work replacement*

| Dimension | Pathway A: extractive replacement | Pathway B: flourishing-centred transition |
|---|---|---|
| Financial security | Rapid redundancy; gains accrue mainly to shareholders; volatile contractor work replaces stable income. | Staged transition; wage insurance; profit-sharing or employee equity; portable benefits and social dividend mechanisms. |
| Agency and dignity | Workers are monitored by AI, given little voice, and informed after deployment decisions are fixed. | Workers participate in task redesign, can contest automated decisions, and retain meaningful human authority. |
| Time and work quality | Remaining roles intensify; availability expectations expand because AI operates continuously. | Productivity gains fund shorter working time, predictable schedules, and removal of low-value administrative burden. |
| Relational well-being | Teams fragment; customer contact is automated even where human reciprocity matters. | Human time is redirected toward mentorship, care, trust-building, and complex interpersonal work. |
| Meaning and development | Occupational identity and progression collapse without replacement pathways. | Learning accounts, civic or creative sabbaticals, new contribution pathways, and recognition beyond billable output are established. |
| Distribution and power | Model ownership and data control concentrate; labour bargaining power falls. | Ownership, data rights, governance, and gains are broadened across affected stakeholders. |
| Planetary impact | Efficiency gains induce more compute-intensive volume without a carbon or resource budget. | Deployment operates within energy, carbon, water, and hardware-lifecycle constraints. |

Pathway A may show higher short-run profit if it externalises transition costs. Pathway B may require greater upfront investment. Yet the causal value profile can favour Pathway B because it converts productivity into security, agency, time, relationships, and durable capability. RoF then asks whether the additional flourishing achieved by Pathway B justifies the incremental transition resources.

The example demonstrates why work replacement cannot be evaluated at the model level or by job counts alone. The proper unit is the **AI-plus-institution transition system**. Evaluation must also be longitudinal. Immediate relief from routine tasks can coexist with later deskilling; an initial income payment can coexist with long-run loss of agency; short-term satisfaction can coexist with weakened relationships. A serious valuation therefore combines baseline comparison, abstention or phased-control groups where feasible, disaggregated outcomes, and follow-up over multiple horizons.

# 7. Institutional Architecture

## 7.1 Firms and investors: dual value accounting

Firms should report a **dual account**: conventional financial performance and a Flourishing Value Account. The latter would disclose the value profile, affected populations, baseline, causal evidence, distribution of gains and harms, threshold status, material uncertainties, and risk-adjusted RoF. This is not a request to monetise every human good. Its purpose is to prevent unpriced outcomes from disappearing from decision processes.

Investment committees could require major AI deployments to pass four gates:

1. **Viability:** Is the intervention financially and operationally feasible?
2. **Additionality:** What changes relative to a credible no-deployment or alternative-deployment baseline?
3. **Flourishing:** Which dimensions and groups improve or deteriorate, over what horizon?
4. **Constraints:** Are agency, distributional, rights, safety, and planetary floors satisfied under downside scenarios?

An intervention can be financially attractive yet fail the flourishing or constraint gates. Conversely, public or philanthropic capital may support high-Flourishing-Value interventions whose benefits are not privately appropriable.

## 7.2 Criteria for a flourishing business through the AI-adoption journey

The net-positive business concept developed by Ruan and Bremen (2025) distinguishes firms that optimise profit alone from those that create societal value incidentally or align strategy deliberately with positive societal outcomes. FVT strengthens that practical idea into a stage-gated definition for AI adoption. A flourishing business does not merely deploy responsible AI or report selected benefits. It organises purpose, design, transition, governance, ownership, and reinvestment so that AI produces durable, fairly distributed flourishing while remaining financially viable.

The following criteria apply across opportunity selection, design and procurement, deployment, operation, scaling, and renewal:

1. **Purpose and success criterion:** the board defines the human and societal purpose of AI adoption and treats durable flourishing, not adoption volume, labour removed, engagement, or profit alone, as the ultimate success criterion.
2. **Materiality, boundary, and stakeholder standing:** the business identifies all materially affected groups, including employees, contractors, customers, non-users, suppliers, communities, future generations, and affected ecosystems, and sets a system boundary wide enough to prevent exported harm from disappearing.
3. **Counterfactual business case:** each material use case states the no-adoption and best-alternative baselines, expected additional value, displaced activities, rebound effects, opportunity costs, distribution of gains and losses, and the evidence that would falsify the investment thesis.

4. **Human capability and complementarity by design:** AI expands effective human capability, judgment, learning, creativity, care, and time sovereignty. Automation is chosen where it removes harmful or unwanted burden; augmentation is preferred where human responsibility, relationship, expertise, or development is constitutive of value.
5. **Just workforce transition:** affected workers participate before decisions are fixed and receive timely notice, retraining, redeployment, income and benefit protection, meaningful appeal, and credible pathways to new forms of contribution. Productivity gains are assessed against work quality, dignity, health, identity, and long-run skill.
6. **Agency, data dignity, and non-manipulation:** people understand when and how AI affects them, can give or withhold meaningful consent where appropriate, can contest consequential decisions, retain human appeal and exit, and are protected from deceptive personalisation, dependency engineering, coercive surveillance, and uncompensated data extraction.
7. **Fair distribution and shared gains:** the allocation of productivity, rents, ownership, bargaining power, time savings, and service improvements is explicit. Workers, customers, data contributors, communities, and public institutions receive a defensible share where their capabilities or resources create the gain.
8. **Safety, security, resilience, and reversibility:** deployment satisfies technical safety, cybersecurity, privacy, robustness, legal, and human-rights requirements; addresses correlated and tail risks; maintains human authority and fallback capacity; and uses staged, monitored, and reversible scaling proportional to consequence.
9. **Relational, civic, meaning, and collective-consciousness effects**: the business tests whether AI strengthens or weakens trust, belonging, reciprocity, public knowledge, social connection, professional purpose, democratic participation, perspective-taking, shared meaning, and the non-coercive capacity to coordinate around common goods. It protects pluralism, dissent, and freedom of conscience, including against slow-moving effects that conventional incident or productivity metrics miss.
10. **Planetary and full-lifecycle integrity:** compute, energy, water, minerals, hardware, waste, biodiversity, and rebound effects are included across the value chain. Efficiency gains do not qualify as flourishing value if scaling increases absolute ecological pressure beyond legitimate boundaries.
11. **Measurement, assurance, and dynamic governance:** the organisation maintains a disaggregated Flourishing Value Account with baselines, causal hypotheses, indicators, affected-group results, uncertainty, threshold status, incidents, and risk-adjusted RoF. Independent assurance, stakeholder review, model and policy updates, and clear redesign or withdrawal triggers operate throughout the lifecycle, consistent with Bremen's (2024, 2025) emphasis on dynamic governance and human-focused metrics.
12. **Net positivity, regeneration, and reinvestment:** after harms, externalities, distribution, and downside risk are internalised, the AI-plus-institution system creates positive societal flourishing value without breaching any floor. A flourishing business then reinvests part of its AI-enabled surplus in human, relational, institutional, knowledge, and natural capacities so that future flourishing capability grows rather than being depleted.

These criteria operate as cumulative gates rather than a menu. Financial viability without positive societal value is insufficient; a positive aggregate score with a breached floor is impermissible; and stated purpose without causal evidence, governance, and affected-group outcomes is not net positivity.

The AI-adoption journey can be represented as four maturity stages. A traditional business optimises AI primarily for financial performance and leaves societal effects outside the decision boundary. An aware or by-product business measures some benefits and harms but does not allow them to govern strategy. A strategic net-positive business aligns investment and governance with positive societal value (Ruan and Bremen 2025). A regenerative flourishing business goes further: it embeds FVT in capital allocation, product and work design, benefit-sharing, assurance, and reinvestment, and expands the durable capability stocks on which future value creation depends.

## 7.3 Governments: from national output to national capability

Post-AGI national accounting should supplement GDP with stocks and flows of flourishing capability. Relevant stocks include population health, financial resilience, skills and learning capacity, relationship and community strength, institutional trust, public knowledge, ecosystem integrity, and widely distributed claims on AI-enabled capital. Relevant flows include changes in each stock, their distribution, and exposure to tail risks.

Governments would use these accounts in fiscal appraisal, infrastructure selection, AI procurement, labour-transition policy, education, health, competition policy, and the public-finance choices created by taxes on labour-replacing capital, compute, tokens, and concentrated AI rents (Guerreiro, Rebelo and Teles 2022; Korinek 2024, 2026). Public value cannot be inferred from aggregate growth when ownership is concentrated or when human and natural capital are depleted. Taxes, social dividends, public equity stakes, sovereign AI funds, data trusts, universal basic services, reduced working time, and predistribution mechanisms should be evaluated as alternative institutional technologies for converting AI rents into flourishing. FVT does not prescribe one mechanism; it supplies the outcome architecture for comparing them.

## 7.4 GDP after AGI: from production accounting to flourishing value accounts

GDP is a production account, not a theory of value. It remains indispensable for measuring market activity, productivity, taxable capacity, debt sustainability, cyclical conditions, and the resources available for public action. Flourishing Value Theory (FVT) therefore does not propose abolishing GDP. It rejects the stronger, often tacit claim that an increase in market-valued output is sufficient evidence that an economy has created societal value. The distinction between production and well-being is already established in the work of the Commission on the Measurement of Economic Performance and Social Progress (Stiglitz, Sen and Fitoussi 2009). Post-AGI conditions make that distinction constitutive of economic analysis rather than a secondary statistical qualification.

Several limitations become more consequential as machine intelligence grows abundant. First, zero-price and non-rival AI outputs, rapid quality change, open knowledge, and AI-enabled process redesign can generate large benefits with a weak or delayed GDP signal. Consumer-surplus extensions such as GDP-B and new approaches to measuring AI improve this boundary, but they do not by themselves establish whose capabilities expanded or whether the gain is durable (Brynjolfsson, Collis and Eggers 2019; Brynjolfsson et al. 2025; Coyle and Poquiz 2025). Second, GDP records expenditure without determining whether the resulting state is worth sustaining. Defensive spending, remediation after cyber or ecological harm, surveillance, and the repair of social breakdown may add to measured output, while unpaid care, trust, civic participation, shared knowledge, spiritual development, and healthy relationships remain weakly represented unless they are monetised. A shift from household or community provision to a paid platform can therefore raise GDP without increasing flourishing.

Third, GDP is distribution-blind. The same aggregate can coexist with radically different ownership of AI capital, access to capability, bargaining power, income security, and exposure to harm. This becomes critical if production separates from human labour: output may accelerate while wages, employment security, social status, meaningful participation, and the institutions through which people claim a share of production deteriorate (Korinek and Stiglitz 2019). Fourth, GDP is primarily a flow measure. It can rise while drawing down the human, relational, institutional, epistemic, and natural-capital stocks that make future flourishing possible. Fifth, it is poorly suited to rights, thresholds, tail risks, irreversibility, and future generations. Preference-shaping AI adds a further problem: systems may increase measured consumption or willingness to pay by manipulating attention and desire while weakening autonomous welfare. No improvement in average output can compensate for the loss of basic agency, severe minority harm, or irreversible planetary damage (Sen 1985, 1999; Nussbaum 2000; Richardson et al. 2023).

The implication of FVT is a layered national-accounting architecture, not the replacement of GDP with another universal scalar. The first layer retains GDP and conventional production, income, financial, and fiscal accounts for the questions they answer well. The second is a National Flourishing Value Profile that reports stocks and flows across physical, emotional, financial, relational, spiritual, and planetary dimensions, together with agency, epistemic capacity, and the collective capacity to learn and govern intelligence wisely. Collective expansion of consciousness belongs here only in an agency-preserving form: reflective awareness, perspective-taking, shared meaning, and non-coercive coordination, protected by pluralism, privacy, dissent, and freedom of conscience.

The third layer is a distribution-and-threshold account. It shows outcomes by affected population, geography, generation, and initial level of capability, and tests whether rights, basic-flourishing floors, or planetary boundaries have been breached. The fourth is an AI Conversion Account. It traces whether AI capability, investment, and rents causally become durable improvements in flourishing, using baselines, additionality, displacement, externalities, time horizon, and Return on Flourishing (RoF). This layer distinguishes the existence of technical abundance from the institutional achievement of flourishing. The fifth is a risk and regenerative balance sheet recording tail exposure, irreversibility, resilience, and changes in the underlying stocks of human, relational, institutional, knowledge, and natural capital.

These layers should remain disaggregated. A headline flourishing index may be useful for communication, but it must be secondary, context-bound, and auditable; it cannot erase distribution, non-substitutable dimensions, rights, or floors. FVT therefore changes the interpretation of GDP growth. Growth is neither presumed beneficial nor treated as irrelevant. It is a production signal whose societal value depends on conversion: who receives the gains, which capabilities expand, which stocks are regenerated or depleted, what risks are created, and whether affected people retain genuine agency. A rise in GDP with negative net societal value is not progress under FVT; a modest GDP path that delivers larger, fairer, and more durable capability gains may be superior.

In practice, national budgets, AI strategies, infrastructure decisions, public procurement, tax design, social dividends, public-equity stakes, and sovereign AI funds should report expected GDP effects beside the Flourishing Value Profile, distribution and threshold status, downside risk, and RoF. GDP would remain a feasibility and macroeconomic-management variable, while public success would be judged by the conversion of productive abundance into health, security, agency, meaningful activity, strong relationships, trustworthy institutions, ecological renewal, and collective capacity to steward intelligence. The post-AGI policy objective is therefore not maximum output. It is durable, fairly distributed flourishing within social and planetary constraints, supported by sufficient and resilient productive capacity.

## 7.5 AI governance: flourishing as system-level observability

Technical safety, security, robustness, fairness, and controllability remain necessary. FVT adds a layer of system-level observability: whether capable systems are producing sustained human and planetary benefit. Developers and regulators should require pre-deployment hypotheses, deployment baselines, post-deployment monitoring, incident triggers, and independent audits across material flourishing dimensions, extending impact-evaluation approaches that separate factual effects from normative judgment (Jenkins et al. 2023) and treating alignment as a normative architecture of human agency rather than a purely technical target (Josifović and Noller 2026).

This role is particularly important for slow-moving harms. Declines in trust, agency, skill, social participation, or meaning may not trigger conventional safety incidents. They can nevertheless indicate deep objective misspecification. Flourishing indicators can function as leading signals of societal misalignment, provided they remain disaggregated and protected from manipulation.

FVT should not be directly installed as a single model reward. Economic models of transformative-technology adoption and catastrophic risk reinforce the case for gradual, reversible deployment and explicit treatment of harms that private actors may not internalise (Acemoglu and Lensman 2024; Jones 2024). The safer architecture is layered: hard safety and rights constraints; plural flourishing indicators; human and democratic governance over weights and thresholds; uncertainty and tail-risk reporting; independent monitoring; and reversible deployment. The more consequential the system, the less legitimate it is for the developer alone to define the value function.

### 7.6 Markets after AGI

Markets will remain useful wherever scarcity, dispersed information, and plural preferences persist. FVT is not a blueprint for replacing markets with central optimisation. It changes the boundary conditions under which market outcomes are judged. Prices can allocate residual scarcities while flourishing accounts evaluate externalities, distribution, public goods, preference integrity, and long-run capability. Democratic institutions can set floors and rights; markets, public provision, commons, and community governance can then operate within those conditions, including commons-oriented alternatives to concentrated AI ownership (Verdegem 2024).

AI agents may reduce transaction costs and expand the market-design frontier, while upstream concentration and delegated consumer choice create new competition and accountability problems (Athey and Scott Morton 2025; Hadfield and Koh 2025; Korinek and Vipra 2025; Shahidi 2025). But lower transaction costs do not guarantee valuable ends. A frictionless market can allocate the wrong commodity space, optimise manipulated preferences, omit relational or ecological externalities, or amplify unequal endowments. Post-AGI economics therefore requires both better coordination mechanisms and an explicit theory of what coordination is for.

## 8. Empirical Research Agenda

FVT becomes scientifically useful only if its claims are testable. Six research programmes follow.

First, researchers should develop **causal flourishing accounts** for AI deployment, using phased rollouts, genuine abstention arms, natural experiments, difference-in-differences designs, and longitudinal follow-up. Correlation between adoption and well-being is insufficient.

Second, measurement work should estimate the relationship between subjective flourishing, behavioural indicators, administrative records, and contextual variables at decision-relevant cadence. Real-time proxies must be validated against slower, richer measures.

Third, work-replacement studies should measure more than employment and wages. Outcomes should include time use, physical and emotional health, financial resilience, agency, learning, dignity, relationships, meaning, community participation, ownership, and ecological burden.

Fourth, research should compare aggregation rules: equal weights, deliberatively chosen weights, prioritarian weights, capability floors, and non-compensatory multi-criteria methods. Results should test sensitivity rather than conceal it.

Fifth, **Flourishing-at-Risk** methods should be developed for advanced AI. This includes scenario distributions, tail dependence across flourishing dimensions, irreversibility, concentration of losses, and institutional risk appetite.

Sixth, governance experiments should compare who specifies value. Developer-defined, customer-defined, worker-participatory, citizen-deliberative, expert, and hybrid models may produce different priorities and legitimacy. The specification process is itself an empirical object.

Business pilots are a natural starting point because organisations can identify a bounded intervention, workforce, baseline, cost, and measurement period. Healthcare, education, digital platforms, and public procurement are also promising. Early pilots should publish the full value profile and sensitivity analysis rather than advertise a single RoF number.

# 9. Objections and Limitations

## 9.1 Is flourishing too paternalistic?

Any substantive welfare theory risks imposing a contested vision of the good. FVT limits this risk through an extensible rather than exhaustive domain structure, direct measurement of subjective experience, capability and agency protections, local and cultural specification, democratic procedures, and transparency about weights. Pure preference satisfaction is not neutral either: it privileges existing endowments, adaptive preferences, and the institutions that shape choice. The relevant comparison is between explicit, contestable value judgments and implicit, privately controlled ones.

## 9.2 Can multidimensional values be compared?

No aggregation rule eliminates ethical judgment. FVT improves comparison by exposing it. Decision-makers see which groups, dimensions, horizons, and assumptions drive a result. Robust decisions can be identified where alternatives rank similarly across plausible weights; contested decisions remain visibly contested. Partial ordering is an acceptable outcome when evidence or moral agreement is insufficient.

## 9.3 Will metrics be Goodharted?

Yes, if treated as targets without safeguards (Goodhart 1975; Manheim and Garrabrant 2019). FVT reduces but cannot eliminate proxy failure. It requires mixed methods, rotating and independently governed indicators, causal validation, adversarial audits, anti-gaming controls, protected qualitative evidence, and retention of the disaggregated profile. Metrics should support judgment and observability, not replace them.

## 9.4 Does the theory understate residual scarcity?

Post-AGI is not post-materiality. Energy, compute, minerals, land, time, attention, biological constraints, ecological sinks, and institutional capacity remain scarce. FVT incorporates these in costs, feasibility conditions, planetary limits, and risk. Its claim is not that scarcity disappears, but that productive intelligence may cease to be the dominant scarcity and that social translation becomes more important.

## 9.5 What about artificial flourishing?

If artificial systems become credible welfare subjects, a human-only value function would be incomplete. The theory therefore treats the welfare-bearing set as revisable through a separate, precautionary moral-status process. Strategic autonomy alone is not sufficient evidence of sentience, and current systems should not receive moral weight merely because they act as market agents. This unresolved boundary is a limitation shared by any post-AGI welfare theory.

### 9.6 Can the framework be captured politically?

Weights, indicators, and thresholds can be captured by states, firms, experts, or majorities. Institutional design is therefore part of validity. Independent measurement bodies, public methods, minority protections, audit rights, plural data sources, appeal, sunset clauses, and decentralised experimentation are necessary. FVT supplies no automatic substitute for constitutional politics.

## 10. Conclusion

The economic theories inherited from industrial and early information economies remain powerful within their domains. Labour explains production and distribution under labour dependence. Prices coordinate residual scarcity. Preferences protect pluralism. Profits sustain investment. GDP tracks market activity. The error is to convert any one of these partial functions into a complete theory of value for an economy transformed by general machine intelligence.

Post-AGI economics begins when producing more intelligence and output is no longer the central civilisational problem. Its central problem is conversion: whether institutions transform that abundance into substantive individual and collective capabilities, and whether those capabilities become health, security, agency, meaningful activity, strong relationships, trustworthy institutions, ecological renewal, and the genuine freedom to pursue different forms of a good life.

Flourishing Value Theory offers a framework for that conversion. It defines value causally rather than transactionally, makes distribution and agency explicit, treats critical dimensions as non-fungible below thresholds, and recognises regenerative and risk-adjusted value across time. At organisational and societal scales, it therefore makes net societal value - the counterfactual contribution of economic activity to fairly distributed flourishing after harms and risks are internalised - the core object of value creation (Ruan and Bremen 2025). It leaves price, profit, and output in place but deprives them of a claim they cannot bear: the claim to tell us, by themselves, whether progress is worth having.

The ultimate post-AGI question is not how much intelligence an economy can deploy. It is what that intelligence enables life to become, and whether it enlarges the individual and collective consciousness required to recognise interdependence, choose worthy ends, and govern intelligence in service of shared flourishing.

**AI-assisted manuscript preparation:** during the preparation of this manuscript, Large Language Models (LLMs) were used to support editorial refinement, formatting, and preparation of submission materials. All ideas, arguments, analyses, interpretations, content and wording were conceived, reviewed, and approved by the author, who accepted full responsibility for the accuracy, originality, and integrity of the work.